# hdlib 2.0: Extending Machine Learning Capabilities of Vector-Symbolic Architectures

Fabio Cumbo [1,*], Kabir Dhillon [1,2], Daniel Blankenberg [1,3]

[1] Center for Computational Life Sciences, Cleveland Clinic Research, Cleveland Clinic, Cleveland, OH 44195, USA
[2] College of Engineering, Ohio State University, Columbus, OH 43210, USA
[3] Department of Molecular Medicine, Cleveland Clinic Lerner College of Medicine, Case Western Reserve University, Cleveland, OH 44195, USA

* To whom correspondence should be addressed:
Fabio Cumbo[1], Center for Computational Life Sciences, Cleveland Clinic Research, Cleveland Clinic, 9500 Euclid Avenue, NA2, Cleveland, OH 44195, USA.
Email: cumbof@ccf.org

## SUMMARY

Following the initial publication of *hdlib* (Cumbo et al. 2023), a Python library for designing Vector-Symbolic Architectures (VSA), we introduce a major extension that significantly enhances its machine learning capabilities. VSA, also known as Hyperdimensional Computing, is a computing paradigm that represents and processes information using high-dimensional vectors. While the first version of *hdlib* established a robust foundation for creating and manipulating these vectors, this update addresses the growing need for more advanced, data-driven modeling within the VSA framework. Here, we present four extensions: significant enhancements to the existing supervised classification model also enabling feature selection, and a new regression model for predicting continuous variables, a clustering model for unsupervised learning, and a graph-based learning model. Furthermore, we propose the first implementation ever of Quantum Hyperdimensional Computing with quantum-powered arithmetic operations and a new Quantum Machine Learning model for supervised learning (Cumbo, Li, et al. 2025).

*hdlib* remains open-source and available on GitHub at https://github.com/cumbof/hdlib under the MIT license, and distributed through the Python Package Index (*pip install hdlib*) and Conda (*conda install -c conda-forge hdlib*). Documentation and examples of these new features are available on the official Wiki at https://github.com/cumbof/hdlib/wiki.

## STATEMENT OF NEED

The successful application of VSA across diverse scientific domains has created a demand for more sophisticated machine learning models that go beyond basic classification. Researchers now require tools to tackle regression tasks, model complex relationships in structured data like graphs, and better optimize models by identifying the most salient features.

This new version of *hdlib* directly addresses this need. While other libraries provide foundational VSA operations (Simon et al. 2022; Heddes et al. 2023; Kang et al. 2022), *hdlib* now introduces a cohesive toolkit for advanced machine learning that is, to our knowledge, unique in its integration of regression, clustering, graph encoding, and enhanced feature selection within a single, flexible VSA framework. These additions empower researchers to move from rapid prototyping of core VSA concepts to building and evaluating complex machine learning pipelines that are now used in the context of different problems in different scientific domains (Cumbo and Chicco 2025; Cumbo, Truglia, et al. 2025; Joshi et al. 2025; Cumbo et al. 2020; Cumbo, Dhillon, Joshi, Chicco, et al. 2025; Cumbo, Dhillon, Joshi, Raubenolt, et al. 2025; Cumbo, Chicco, et al. 2025).

## EXTENDING MACHINE LEARNING FUNCTIONALITIES

The primary contribution of this work is the expansion of the *hdlib.model* module with new functionalities to enhance existing methods and the introduction of new modules for handling different data structures. The new architecture is summarized in Figure 1.

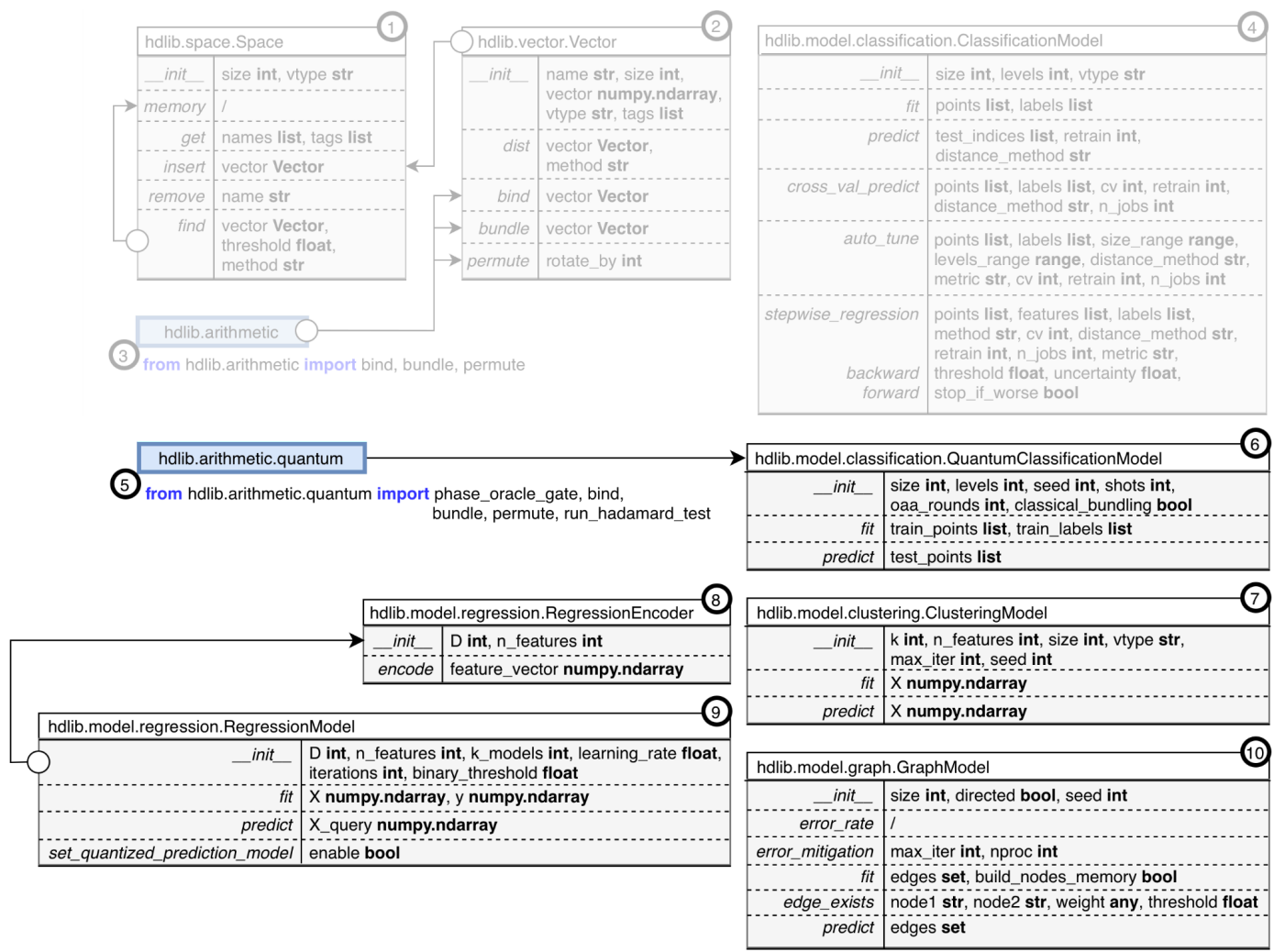

**Figure 1:** An overview of the *hdlib* 2.0 library architecture, highlighting the distinction between the original (top, transparent) and new components (bottom). Foundational classes from version 1.0 include *hdlib.space.Space* (Class 1), *hdlib.vector.Vector* (Class 2), *hdlib.arithmetic* module (Module 3), and the *hdlib.model.classification.ClassificationModel* (Class 4). This work introduces major new models and features summarized with the *hdlib.arithmetic.quantum* (Module 5) module and the

> *hdlib.model.classification.QuantumClassificationModel* (Class 6), and the *hdlib.model* module which comprises the new *clustering.ClusteringModel* (Class 7), *regression.RegressionEncoder* (Class 8) and *regression.RegressionModel* (Class 9), and *graph.GraphModel* (Class 10), creating a comprehensive toolkit for VSA-based machine learning.

## Classification Model

A key focus of this update was to provide more robust and automated tools for model optimization:

- **Enhanced feature selection:** the original *hdlib.model.Model* class (now *hdlib.model.classification.ClassificationModel*) provided a *stepwise_regression* instance method for feature selection. This functionality has been significantly enhanced to offer greater control over the selection process, improved performance, and more detailed reporting on feature importance. This refinement helps in building more interpretable VSA models;
- **Advanced hyperparameter tuning:** the initial version of the library included an *auto_tune* instance method for performing a parameter sweep analysis on vector dimensionality and the number of level vectors. This has been upgraded to a more advanced hyperparameter optimization tool. The new implementation is more efficient and allows for a more thorough and effective search of the hyperparameter space to automatically maximize the model performances.

## Clustering Model

Here, we introduced a new *hdlib.model.clustering* module that provides a *ClusteringModel* class that implements a k-means clustering algorithm working accordingly with the Hyperdimensional Computing principles as defined in (Gupta et al. 2022).

The algorithm operates by representing both the *k* cluster centroids and the input data points as hypervectors. The iterative *fit* process closely mirrors the classic k-means algorithm but uses VSA operations. In the assignment step of each iteration, data points are assigned to the cluster corresponding to the most similar centroid, determined by calculating the cosine distance in the high-dimensional space. In the subsequent update step, the centroid of each cluster is recalculated by performing a bundling operation (element-wise addition and normalization) on all the hypervectors of the data points assigned to it. This process naturally moves the centroid towards the center of its constituent points. This iterative process continues until the cluster assignments stabilize or a maximum number of iterations is reached. Once the model is trained, the *predict* method can be used to assign a new, unseen data point to the most appropriate cluster.

## Regression Model

To address tasks involving the prediction of continuous variables, *hdlib* now implements a regression model based on the methodology described by (Hernández-Cano et al. 2021). This implementation is split into two main components: a *RegressionEncoder* and a *RegressionModel* as part of the *hdlib.model.regression* module. The encoder maps input features into a high-dimensional space using a non-linear function that combines the input with a set of random base hypervectors and biases. This mapping is specifically designed to preserve the similarity relationships of the original feature space.

The *RegressionModel* employs a sophisticated multi-model strategy, maintaining a set of *k* parallel cluster models and regression models. During the iterative *fit* process, an encoded input vector is compared against all cluster models to compute a set of confidence scores via a softmax function. A final prediction is produced by a confidence-weighted sum of the outputs from all regression models. The prediction error is then used to update the models: all regression models are adjusted based on their confidence score, while only the most similar cluster model is refined. This process allows the system to learn complex, non-linear relationships in the data. For efficiency, the module can maintain both full-precision and binarized versions of the models, and users can enable a *quantized_prediction* mode for accelerated inference using Hamming distance. This enables VSA to be applied to a new class of problems, such as predicting physical properties, financial values, or other scalar quantities.

**Graph Model**

A major extension in this release is the *hdlib.model.graph* module, which provides the *GraphModel* class for representing and reasoning with graph-based data. This implementation encodes an entire directed and undirected weighted graph into a single hyperdimensional vector, based on the methodology described by (Poduval et al. 2022). The process begins by assigning a unique random hypervector to each node and edge weight. The *fit* method then constructs the graph representation by first creating a memory vector for each node that encodes its local neighborhood. This is achieved by bundling the vectors of its neighbors, each binded with their respective edge-weight vector. Finally, the entire graph is compressed into one vector by bundling all node vectors, each binded with its corresponding memory vector. For directed graphs, a *permute* operation is used to preserve the directionality of edges within the final representation.

Crucially, the library can query the existence of an edge directly from this single graph vector. The *edge_exists* method uses binding operations to probe the graph vector, retrieve a noisy version of a node's memory, and check its similarity to a potential neighbor. Furthermore, the module includes a *predict* method for edge weight classification and an *error_mitigation* routine for iteratively refining the graph model to reduce prediction errors, making it a complete toolkit for graph-based machine learning.

With the integration of these modules, *hdlib* 2.0 provides the scientific community with a unified and powerful framework, paving the way for the development of novel, brain-inspired solutions to a broader spectrum of machine learning problems.

**Quantum Hyperdimensional Computing**

We introduce Quantum Hyperdimensional Computing (QHDC), a foundational paradigm designed to run on quantum devices (Cumbo, Li, et al. 2025). The *hdlib.arithmetic.quantum* module implements the QHDC arithmetic using IBM's Qiskit framework (Javadi-Abhari et al. 2024):

- **Encoding:** we employ a phase encoding strategy (Khan et al. 2024) where bipolar hypervectors are mapped to the relative phases of a uniform superposition state, enabling efficient algebraic manipulation;
- **Binding:** realized via quantum phase oracles that map element-wise multiplication to the sequential application of phases (Chen et al. 2024);
- **Bundling:** implemented as a quantum-native averaging process using a Linear Combination of Unitaries (LCU) (Chakraborty 2024) followed by Oblivious Amplitude Amplification (OAA) (Guerreschi 2019);
- **Permutation:** achieved using the Quantum Fourier Transform (QFT) (Weinstein et al. 2001) to induce cyclic shifts in the computational basis;
- **Similarity:** computed via the Hadamard Test (Mehta et al. 2025) to estimate the real part of the inner product (cosine similarity) between quantum states.

Finally, we provide a *QuantumClassificationModel* class for supervised learning as a totally new approach to Quantum Machine Learning (Biamonte et al. 2017).

With the integration of these modules, *hdlib* 2.0 provides the scientific community with a unified and powerful framework, paving the way for the development of novel, brain-inspired solutions to a broader spectrum of machine learning problems.

**REFERENCES**

Biamonte, Jacob, Peter Wittek, Nicola Pancotti, Patrick Rebentrost, Nathan Wiebe, and Seth Lloyd. 2017. "Quantum Machine Learning." *Nature* 549 (7671): 195–202.

Chakraborty, Shantanav. 2024. "Implementing Any Linear Combination of Unitaries on Intermediate-Term Quantum Computers." *Quantum* 8 (October): 1496.

Chen, X., A. Dua, M. Hermele, et al. 2024. "Sequential Quantum Circuits as Maps between Gapped Phases." *Physical Review B* 109 (February). https://doi.org/10.1103/PhysRevB.109.075116.

Cumbo, Fabio, Eleonora Cappelli, and Emanuel Weitschek. 2020. "A Brain-Inspired Hyperdimensional Computing Approach for Classifying Massive DNA Methylation Data of Cancer." *Algorithms* 13 (9): 233.

Cumbo, Fabio, and Davide Chicco. 2025. "Hyperdimensional Computing in Biomedical Sciences: A Brief Review." *PeerJ. Computer Science* 11 (May): e2885.

Cumbo, Fabio, Davide Chicco, Sercan Aygun, and Daniel Blankenberg. 2025. "Designing Vector-Symbolic Architectures for Biomedical Applications: Ten Tips and Common Pitfalls." In *Preprints.org*. October 2. https://doi.org/10.20944/preprints202510.0117.v1.

Cumbo, Fabio, Kabir Dhillon, Jayadev Joshi, Bryan Raubenolt, et al. 2025. "Predicting the Toxicity of Chemical Compounds via Hyperdimensional Computing." In *bioRxiv*. September 21. https://doi.org/10.1101/2025.09.12.675894.

Cumbo, Fabio, Kabir Dhillon, Jayadev Joshi, Davide Chicco, Sercan Aygun, and Daniel Blankenberg. 2025. "A Novel Vector-Symbolic Architecture for Graph Encoding and Its Application to Viral Pangenome-Based Species Classification." In *bioRxiv*. September 10. https://doi.org/10.1101/2025.09.08.674958.

Cumbo, Fabio, Rui-Hao Li, Bryan Raubenolt, et al. 2025. "Quantum Hyperdimensional Computing: A Foundational Paradigm for Quantum Neuromorphic Architectures." In *arXiv*. November 16. http://arxiv.org/abs/2511.12664.

Cumbo, Fabio, Simone Truglia, Emanuel Weitschek, and Daniel Blankenberg. 2025. "Feature Selection with Vector-Symbolic Architectures: A Case Study on Microbial Profiles of Shotgun Metagenomic Samples of Colorectal Cancer." *Briefings in Bioinformatics* 26 (2). https://doi.org/10.1093/bib/bbaf177.

Cumbo, Fabio, Emanuel Weitschek, and Daniel Blankenberg. 2023. "Hdlib: A Python Library for Designing Vector-Symbolic Architectures." *Journal of Open Source Software* 8 (89): 5704.

Guerreschi, Gian Giacomo. 2019. "Repeat-until-Success Circuits with Fixed-Point Oblivious Amplitude Amplification." *Physical Review A* 99 (February). https://doi.org/10.1103/PhysRevA.99.022306.

Gupta, Saransh, Behnam Khaleghi, Sahand Salamat, et al. 2022. "Store-N-Learn: Classification and Clustering with Hyperdimensional Computing across Flash Hierarchy." *ACM Transactions on Embedded Computing Systems* 21 (3): 1–25.

Heddes, Mike, Igor Nunes, Pere Vergés, et al. 2023. "Torchhd: An Open Source Python Library to Support Research on Hyperdimensional Computing and Vector Symbolic Architectures." *Journal of Machine Learning Research* 24 (255): 1–10.

Hernández-Cano, Alejandro, Cheng Zhuo, Xunzhao Yin, and Mohsen Imani. 2021. "RegHD: Robust and Efficient Regression in Hyper-Dimensional Learning System." Paper presented 2021 58th ACM/IEEE Design Automation Conference (DAC). November 8. https://doi.org/10.1109/DAC18074.2021.9586284.

Javadi-Abhari, Ali, Matthew Treinish, Kevin Krsulich, et al. 2024. "Quantum Computing with Qiskit." In *arXiv*. May 14. http://arxiv.org/abs/2405.08810.

Joshi, Jayadev, Fabio Cumbo, and Daniel Blankenberg. 2025. "Large-Scale Classification of Metagenomic Samples: A Comparative Analysis of Classical Machine Learning Techniques vs a Novel Brain-Inspired Hyperdimensional Computing Approach." In *bioRxiv*. July 7. https://doi.org/10.1101/2025.07.06.663394.

Kang, Jaeyoung, Behnam Khaleghi, Tajana Rosing, and Yeseong Kim. 2022. "OpenHD: A GPU-Powered Framework for Hyperdimensional Computing." *IEEE Transactions on Computers* 71 (11): 2753–2765.

Khan, M. A., M. N. Aman, and B. Sikdar. 2024. "Beyond Bits: A Review of Quantum Embedding Techniques for Efficient Information Processing." *IEEE Access* 12 (March). https://doi.org/10.1109/ACCESS.2024.3382150.

Mehta, Vivek, Arghya Choudhury, and Utpal Roy. 2025. "Generalized Quantum Hadamard Test for Machine Learning." In *arXiv*. August 6. http://arxiv.org/abs/2508.04065.

Poduval, Prathyush, Haleh Alimohamadi, Ali Zakeri, et al. 2022. "GrapHD: Graph-Based Hyperdimensional Memorization for Brain-Like Cognitive Learning." *Frontiers in Neuroscience* 16 (February): 757125.

Simon, William Andrew, Una Pale, Tomas Teijeiro, and David Atienza. 2022. "HDTorch: Accelerating Hyperdimensional Computing with GP-GPUs for Design Space Exploration." Paper presented ICCAD '22. *ICCAD '22: Proceedings of the 41st IEEE/ACM International Conference on Computer-Aided Design*, December 22. https://doi.org/10.1145/3508352.3549475.

Weinstein, Y. S., M. A. Pravia, E. M. Fortunato, S. Lloyd, and D. G. Cory. 2001. "Implementation of the Quantum Fourier Transform." *Physical Review Letters* 86 (February). https://doi.org/10.1103/PhysRevLett.86.1889.